\documentclass[11pt]{academic_template}
\usepackage[utf8]{inputenc}
\usepackage[T1]{fontenc}
\usepackage{lmodern}
\usepackage{amsmath,amssymb,amsfonts,bm}
\usepackage{nicefrac,enumitem}
\usepackage{algorithm,algpseudocode}
\usepackage{wrapfig,listings}
\usepackage{tabularx,array,float}
\usepackage{graphicx,xcolor}
\usepackage{subcaption}
\setthemecolor{2E5AA8}

\newcommand{\KL}{\mathrm{KL}}
\newcommand{\E}{\mathbb{E}}
\newcommand{\base}{\pi_{0}}
\newcommand{\teacher}{\pi^{\mathrm{T}}}
\newcommand{\student}{\pi_{\theta}}

\title{VLA-MOPD: On-Policy Distillation for Multi-Task VLA Policy Consolidation}
\author[1,2,*]{Yongxuan Lv}
\author[1,*]{Zhen Fang}
\author[2,3,*]{Lichen Man}
\author[1]{Hao Li}
\author[2]{Pu Li}
\author[1,2]{Hengyi Wang}
\author[1]{Langfeng Cao}
\author[1]{Lin Chen}
\author[1]{Jie Huang}
\author[2,3]{Xiu Su}
\author[2]{Shan You}
\author[2]{Fei Wang}
\author[1,\dagger]{Feng Zhao}
\affiliation[1]{University of Science and Technology of China}
\affiliation[2]{ACE Robotics}
\affiliation[3]{Central South University}
\contribution[*]{Equal contribution}
\contribution[\dagger]{Corresponding author}
\abstract{%
Physical intelligence requires embodied agents to ground language in observations and execute actions while retaining useful skills. Vision-language-action (VLA) policies offer a shared interface, but task-specific post-training often produces separate specialists. Consolidation is difficult: mixed-task reinforcement learning provides sparse trajectory-level credit, while offline distillation may miss student-visited observations and endpoint distillation does not supervise intermediate action-generation states. We introduce VLA Multi-teacher On-Policy Distillation (VLA-MOPD), a protocol for consolidating flow-matching VLA specialists. After acquiring task or suite experts, a shared student collects robot rollouts and queries the instruction-routed frozen teacher at each visited observation and intermediate ordinary differential equation (ODE) state. Detached per-flow-step transition targets supervise valid action chunks through a masked local matching loss; an optional reference-policy term supports retention on other tasks. Starting from one trajectory per LIBERO task, VLA-MOPD raises mean success-once from 75.4\% to 98.9\%, compared with 98.15\% for endpoint action distillation. It improves mean RoboTwin 2.0 success from 73.0\% to 89.3\%; on a Piper robot, two consolidation rounds improve average success from 70.0\% to 90.0\%. In a near-matched Spatial-adaptation study, reference matching limits the Object suite decrease to 3.0 points, versus 13.2 without the reference term.
}

\begin{document}
\maketitle
\begin{center}
\small Code and model links: \url{https://github.com/yxuan1234/VLA-MOPD}
\end{center}
\newpage
\begin{spacing}{0.9}
\tableofcontents
\end{spacing}
\newpage

\section{Introduction}
\label{sec:intro}

Physical intelligence requires agents to connect language and perception to
actions whose consequences change future observations. Embodied agents must
therefore ground instructions in the current physical state and execute actions
with feedback. PaLM-E incorporates visual and robot-state inputs into a
language model for embodied reasoning, while SayCan grounds high-level plans
in the feasibility of pretrained robot skills~\citep{palme,saycan}.

VLA models provide a direct interface from
observations and instructions to robot actions. RT-1 and RT-2 established
large-scale language-conditioned control, while Open X-Embodiment broadened
the available training data across institutions and embodiments
\citep{rt1,rt2,openx}. Octo and OpenVLA provide reusable generalist
initializations~\citep{octo,openvla}. Diffusion- and flow-based policies
generate continuous action chunks through an iterative trajectory
\citep{diffusionpolicy,pi0,pi05}. Yet task-specific post-training often leaves
improvements distributed across separate specialist checkpoints.

We study how to consolidate these specialists into one flow-matching VLA. A
flow-matching VLA encounters two coupled trajectories: the \emph{outer robot
trajectory}, whose states are observations reached after executing action
chunks, and the \emph{inner flow trajectory}, whose ODE states transform noise
into each action chunk. Fixed specialist demonstrations may miss observations
reached by the changing student. Because closed-loop errors alter future states, the relevant training
distribution is the student's own occupancy. On-policy reinforcement learning (RL) naturally visits
this distribution, but the multi-task Group Relative Policy Optimization
(GRPO) baseline, which we refer to as MixRL~\citep{simplevla,pirl}, assigns one binary success
value to an entire rollout. This gives limited
credit information for individual action chunks, especially before successful
rollouts become common, and shared updates can encounter cross-task gradient
interference~\citep{pcgrad}.

Distillation provides direct teacher targets on student states. Classical
interactive imitation learning such as Dataset Aggregation (DAgger) queries an expert for an action
at each learner-induced observation~\citep{dagger}. In our setting, matching
the teacher's final action at these observations gives the endpoint action
distillation baseline. We instead draw on DiffusionOPD and Flow-OPD, which
study local transition or
vector-field matching in continuous generative models~\citep{diffusionopd,flowopd},
to provide denser supervision: the routed teacher supplies a detached
per-flow-step transition target at every inner flow state. VLA-MOPD first
acquires and freezes task or suite specialists, then lets a student collect
robot rollouts and routes each instruction to its teacher. The student matches
these targets across valid action chunks, including failed rollouts, while an
optional reference term applies the same rule to a frozen copy of the initial
policy on retained tasks.

We evaluate VLA-MOPD on four LIBERO suites and three RoboTwin 2.0 tasks
\citep{libero,robotwin2}, as well as three tasks on a Piper robot. Starting
from one demonstration per LIBERO task, mean success-once rises from $75.4\%$
to $98.9\%$, compared with $98.15\%$ for endpoint action distillation. The
RoboTwin mean rises from $73.0\%$ to $89.3\%$; on the Piper robot, two
consolidation rounds raise average success from $70.0\%$ to $90.0\%$. In a
near-matched Spatial-adaptation comparison, reference matching limits the Object
suite decrease to $3.0$ points, versus $13.2$ without the reference term.

Our contributions are (i) a two-stage recipe that decouples specialist
acquisition from multi-task VLA consolidation; (ii) a local ODE transition
matching objective evaluated at student-induced outer observations and inner
flow states; and (iii) evidence of capability consolidation across simulated
benchmarks and real-robot tasks, together with an analysis of retention.

\section{Related Work}
\label{sec:related}

\subsection{Vision-Language-Action Models}
Scaling robot data and model capacity has produced policies that transfer
across tasks, embodiments, and language instructions. RT-1 and RT-2 study
large-scale real-robot and vision--language policy training, while Open
X-Embodiment standardizes multi-institution robot data for cross-embodiment
learning \citep{rt1,rt2,openx}. Octo and OpenVLA provide open generalist
initializations; OpenVLA-OFT shows that action representation, chunking, and
decoding choices affect downstream adaptation \citep{octo,openvla,openvla_oft}.
Diffusion Policy and the $\pi_0$ family instead generate continuous action
sequences through diffusion or flow matching \citep{diffusionpolicy,pi0,pi05}.
Recent VLA designs explore partial reasoning--action decoupling, instruction
tuning, hybrid diffusion--autoregressive generation, fast--slow control,
latent spatio-temporal reasoning, multi-frame modeling, execution-status
monitoring, adaptive test-time compute, and motion-aware planning
\citep{fang2025dualvla,instructvla,chen2025fast,liu2025hybridvla,liu2026last,li2025cronusvla,sentinelvla,vlaattc,dsdynvla}.
These works provide the policy and adaptation substrate for our study.

\subsection{Interactive and Reinforcement Learning for VLA Post-Training}
Interactive imitation learning addresses distribution shift by querying an
expert on learner-induced observations. DAgger is the canonical example
\citep{dagger}; our setting retains this outer on-policy
idea and adds targets at intermediate states of a continuous action generator.
Recent embodied learning methods use environment interaction as additional
supervision, including agentic reward verification for broader rollout
exploration \citep{rewardasagent}; in VLA post-training, SimpleVLA-RL applies
GRPO to an autoregressive action-token VLA, while $\pi_{\mathrm{RL}}$
introduces Flow-Noise and Flow-SDE variants for Proximal Policy Optimization
(PPO) or GRPO on flow-based
$\pi_0$ and $\pi_{0.5}$ policies \citep{simplevla,pirl}. VLA-OPD queries a frozen teacher for action-token
distributions on student-visited states and optimizes a reverse Kullback--Leibler (KL) objective
\citep{vlaopd}. Residual controllers and lightweight actor--critic heads offer
other adaptation routes \citep{residual_policy,residual_rl,rltoken,pld}.

Most of these methods emphasize task-specific adaptation or direct improvement
from interaction. Our setting first acquires task- or suite-level specialists
and then integrates them into one shared student, emphasizing consolidation
across complementary capabilities.

\subsection{On-Policy Distillation for Multi-Teacher Consolidation}
Building on multi-teacher knowledge distillation \citep{you2017multiple},
on-policy distillation (OPD) separates specialist acquisition from integration by training
domain specialists and distilling them into one student on the student's own
rollouts. Language-model work uses dense token-level supervision for OPD
\citep{opd,mopd,mimo_report,deepseekv4}, while process-reward
policy optimization provides intermediate reasoning-level credit for
multi-step reasoning \citep{prpo}. VLA-OPD applies this idea to an
autoregressive action-token policy and matches token distributions at visited
observations. Our student instead matches continuous flow-action transitions:
the flow-matching action expert exposes intermediate ODE states and step means
that provide the local targets used for consolidation.

MergeVLA instead preserves task-specific components through sparse task masks
and composable action experts, whereas our focus is teacher-guided on-policy
consolidation of a shared flow-matching policy \citep{mergevla}.

For continuous generative models, DiffusionOPD derives a closed-form per-step
KL objective for diffusion transitions. Flow-OPD adapts the same idea to flow
matching: it routes student samples to the relevant specialist and uses
vector-field discrepancies as dense feedback for policy-gradient updates
\citep{diffusionopd,flowopd}. We transfer this transition-matching recipe to
flow-matching VLA action experts: the outer robot rollout supplies visited
observations, the inner flow trajectory supplies intermediate states, and the
task-routed teacher supplies a target at each state.

\section{Motivation: Decoupling Skill Acquisition and Consolidation}
\label{sec:diagnosis}

The consolidation objective must evaluate teacher targets where the student
actually acts and generates action chunks. The following bound isolates the
effect of replacing that joint state distribution with another one.

\paragraph{State-distribution mismatch.}
Let $\ell(c,x)$ be a bounded per-state matching loss and let $d$ and
$\widetilde d$ be two observation--flow-state distributions. We measure their
discrepancy by the total variation (TV) distance. If
$0\leq\ell\leq M$, then
\begin{equation}
\left|\E_{d}[\ell]-\E_{\widetilde d}[\ell]\right|
\leq M\,\mathrm{TV}(d,\widetilde d),
\label{eq:state_shift_bound}
\end{equation}
which bounds how much the mismatch between the two state distributions can
change the objective. This change-of-measure bound formalizes why student-induced states are the
natural evaluation distribution for transition matching.

\section{Methodology}
\label{sec:method}

\subsection{Framework Overview}
\begin{figure}[t]
\centering
\includegraphics[width=\linewidth]{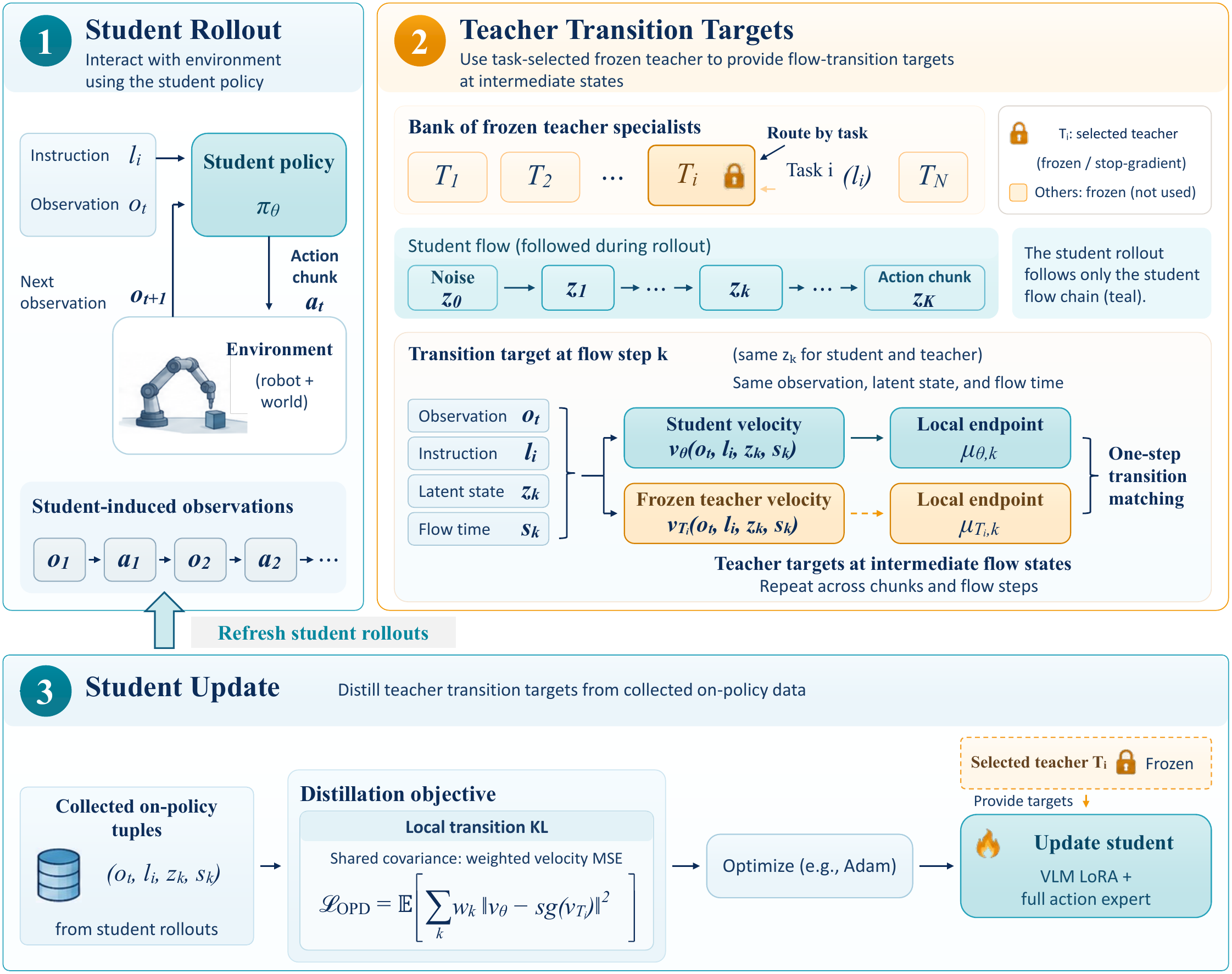}
\caption{VLA-MOPD consolidation. Student rollouts provide observations and ODE
states; the task-routed specialist labels transitions. The student matches
these targets and refreshes rollouts. Velocity and ODE step-mean regression are
Euler-equivalent; the transition-KL interpretation assumes shared Gaussian
covariance (Eq.~\ref{eq:gaussian}).}
\label{fig:method_v3}
\end{figure}

Multi-task supervised fine-tuning (SFT) initializes $\pi_{\theta_0}$;
supervised or RL adaptation produces frozen
specialists $\teacher_i=\pi_{\phi_i}$ for tasks or suites
$i\in\mathcal I$. Figure~\ref{fig:method_v3} shows each consolidation
iteration:
one student rollout collection, task-routed transition labeling, and one
student update. Its
$z_k$ denotes $x_{t,k}$ below, and its local endpoint is the ODE step mean at
flow step $k$.
The trainable student parameters are its vision-language model (VLM)
Low-Rank Adaptation (LoRA) modules and full action expert.

At decision step $t$, let $c_t=(o_t,\ell_i)$ contain the observation and task
instruction. The flow-matching head generates a normalized action chunk in
$\mathbb R^{H\times d}$ from Gaussian noise with an ODE solver. For $1=s_0>\cdots>s_K=0$, define
$\delta_k=s_k-s_{k+1}$ and the ODE step mean at flow step $k$
\begin{equation}
\mu_{\theta,k}(x,c)=x-\delta_k v_\theta(x,s_k,c),
\qquad k=0,\ldots,K-1,
\label{eq:ode_mean}
\end{equation}
where $v_\theta$ is the velocity field and $\mu_{\theta,k}$ is the
\emph{ODE step mean}. A frozen snapshot generates
$x_{t,0}\sim\mathcal N(0,I)$ and recursively applies
$x_{t,k+1}=\mu_{\bar\theta,k}(x_{t,k},c_t)$; $x_{t,K}$ is decoded and executed
for training rollouts. During evaluation, the controller may execute only a
prefix before requesting a new action chunk.
We store $c_t$, the teacher identifier, and the full chain $x_{t,0:K}$.

At these states, the routed teacher supplies detached targets
$\widehat\mu_{i,t,k}=\operatorname{sg}[\mu_{\phi_i,k}(x_{t,k},c_t)]$, where
$\operatorname{sg}[\cdot]$ denotes a stop-gradient.
Teacher and student compute separate VLM prefixes but predict in shared
normalized action coordinates. We cache targets, update only the student's VLM
LoRA and action expert, and refresh the student rollout after each update
(Algorithm~\ref{alg:vmopd}).

\begin{algorithm}[t]
\caption{VLA-MOPD: Specialist Acquisition and On-Policy Consolidation}
\label{alg:vmopd}
\begin{algorithmic}[1]
\Require Multi-task SFT data $\mathcal D_{\mathrm{SFT}}$, specialist acquisition procedure $\mathcal A$
\Require Teacher index set $\mathcal I$, task sampling schedule $p(i)$, and flow schedule $s_{0:K}$
\State $\theta_0\gets\operatorname{SFT}(\mathcal D_{\mathrm{SFT}})$; $\theta\gets\theta_0$
\For{$i\in\mathcal I$}
    \State $\phi_i\gets\mathcal A(i)$; freeze $\phi_i$
\EndFor
\State $\theta_{\mathrm{train}}\gets(\theta_{\mathrm{VLM\text{-}LoRA}},\theta_{\mathrm{action}})$
\While{the configured training budget remains}
    \State $\bar\theta\gets\theta$
    \State $\mathcal D_{\mathrm{roll}}\gets\operatorname{RolloutODE}(\pi_{\bar\theta},p(i),s_{0:K})$
    \Statex \hspace{\algorithmicindent}\hspace{\algorithmicindent}$\mathcal D_{\mathrm{roll}}=\{(i,c,x_{0:K},m)\}$; $m$ keeps valid action chunks and masks terminated or padded rows
    \For{each $(i,c,x_{0:K},m)\in\mathcal D_{\mathrm{roll}}$}
        \State $e_i^{\mathrm T}\gets\operatorname{Prefix}_{\phi_i}(c)$ \Comment{Teacher's own VLM}
        \For{$k=0,\ldots,K-1$}
            \State $v_{i,k}^{\mathrm T}\gets\operatorname{Velocity}_{\phi_i}(x_k,s_k,c;e_i^{\mathrm T})$
            \State $\widehat\mu_{i,k}\gets\operatorname{sg}[x_k-\delta_k v_{i,k}^{\mathrm T}]$; cache $\widehat\mu_{i,k}$
        \EndFor
    \EndFor
    \For{each update minibatch $\mathcal B\subset\mathcal D_{\mathrm{roll}}$}
        \State $e^{\mathrm S}\gets\operatorname{Prefix}_{\theta}(c_{\mathcal B})$ \Comment{Trainable student prefix}
        \For{$k=0,\ldots,K-1$}
            \State $v_{\theta,k}\gets\operatorname{Velocity}_{\theta}(x_{\mathcal B,k},s_k,c_{\mathcal B};e^{\mathrm S})$
            \State $\mu_{\theta,k}\gets x_{\mathcal B,k}-\delta_k v_{\theta,k}$
        \EndFor
        \State $\widehat{\mathcal L}_{\mathrm{ODE}}\gets\operatorname{MaskedMeanMatch}(\mu_\theta,\widehat\mu,m)$ \Comment{Eq.~\ref{eq:ode_loss}}
        \State $\theta_{\mathrm{train}}\gets\operatorname{Update}(\theta_{\mathrm{train}},\nabla_{\theta_{\mathrm{train}}}\widehat{\mathcal L}_{\mathrm{ODE}})$
    \EndFor
\EndWhile
\State \Return $\pi_\theta$
\end{algorithmic}
\end{algorithm}

\subsection{Optimization Objective}
\label{sec:objective}
Following DiffusionOPD \citep{diffusionopd}, we start from a stochastic
transition-KL view, then use direct transition matching for our deterministic
ODE sampler. Let $d_i^{\bar\theta}(c)$
be the student snapshot's observation--instruction distribution and
$q_{\bar\theta}(x_{0:K}\mid c)$ its ODE-chain law. At each stored $x_k$, we
compare the student and frozen-teacher transitions. For a fixed snapshot
$\bar\theta$, the stochastic reference objective is
\begin{equation}
\mathcal L_{\mathrm{OPD}}(\theta;\bar\theta)
=\E_{\substack{i\sim p(i),\ c\sim d_i^{\bar\theta}\\
 x_{0:K}\sim q_{\bar\theta}(\cdot\mid c)}}
\left[\frac{1}{K}\sum_{k=0}^{K-1}
 \KL\!\left(p_{\theta,k}(\cdot\mid x_k,c)
 \,\|\,p^{\mathrm T}_{i,k}(\cdot\mid x_k,c)\right)\right].
\label{eq:local}
\end{equation}
Here the student supplies both the robot observations and intermediate flow
states, while the instruction selects the teacher.

\paragraph{Closed-form KL.}
For the closed-form transition calculation, first consider stochastic transitions with shared
covariance. Write $p_{\theta,k}=\mathcal N(\mu_{\theta,k},\Sigma_k)$ and
$p^{\mathrm T}_{i,k}=\mathcal N(\mu_{\phi_i,k},\Sigma_k)$, with both means
conditioned on the same $(x_k,c)$. Using the Euler parameterization
$\mu_{\theta,k}=x_k-\delta_k v_\theta(x_k,s_k,c)$ and
$\mu_{\phi_i,k}=x_k-\delta_k v_{\phi_i}(x_k,s_k,c)$, the common state cancels and
\begin{equation}
\KL\!\left(p_{\theta,k}\,\|\,p^{\mathrm T}_{i,k}\right)
=\tfrac12\Delta\mu_k^\top\Sigma_k^{-1}\Delta\mu_k
=\frac{\delta_k^2}{2\sigma_k^2}
\|v_\theta(x_k,s_k,c)-v_{\phi_i}(x_k,s_k,c)\|_F^2
\quad\text{if }\Sigma_k=\sigma_k^2 I,
\label{eq:gaussian}
\end{equation}
where $\Delta\mu_k=\mu_{\theta,k}-\mu_{\phi_i,k}$. This is an exact KL only
for the stated stochastic Gaussian kernels.

\paragraph{Deterministic ODE matching.}
Our action generator instead uses deterministic Euler steps
$x_{k+1}=\mu_{\theta,k}(x_k,c)$. As in the deterministic regime of
DiffusionOPD, we directly match the student and teacher ODE step means at the same stored
student state:
\begin{equation}
\ell^{\mathrm{ODE}}_{i,k}(x_k,c)
=\tfrac12\|\mu_{\theta,k}(x_k,c)-\mu_{\phi_i,k}(x_k,c)\|_F^2
=\tfrac{\delta_k^2}{2}\|v_\theta(x_k,s_k,c)-v_{\phi_i}(x_k,s_k,c)\|_F^2.
\label{eq:ode_step_match}
\end{equation}
This is a squared transition-matching surrogate, not a KL between
deterministic kernels: distinct Dirac transitions have no finite KL.

\paragraph{VLA action-chunk objective.}
For each robot observation $c_t$, the student generates an $H\times d$ action
chunk and stores its inner ODE states. We query the routed teacher at those
states, detach its targets, and average Eq.~\ref{eq:ode_step_match} over flow
steps and action elements:
\begin{equation}
\mathcal L_{\mathrm{ODE}}(\theta;\bar\theta)
=\E_{\substack{i\sim p(i),\ c\sim d_i^{\bar\theta}\\
 x_{0:K}\sim q_{\bar\theta}(\cdot\mid c)}}
\left[\frac{1}{2KHd}\sum_{k=0}^{K-1}
 \|\mu_{\theta,k}(x_k,c)
 -\operatorname{sg}[\mu_{\phi_i,k}(x_k,c)]\|_F^2\right].
\label{eq:ode_loss}
\end{equation}
Equation~\ref{eq:ode_loss} uses uniform mean-space weighting, so its
equivalent velocity error carries $\delta_k^2$ but no inverse-covariance
factor. Appendix~\ref{app:implementation} gives masking and normalization
details; Appendix~\ref{app:reverse_kl} expands the KL-to-ODE connection.

\paragraph{Frozen-snapshot projection.}
For a fixed stored state, write
$J_{\theta,k}=\partial v_\theta(x_k,s_k,c)/\partial\theta$. Differentiating the velocity-field objective above gives the update direction
\begin{equation}
\nabla_\theta \mathcal L_{\mathrm{ODE}}
=\E\left[\frac{1}{KHd}\sum_{k=0}^{K-1}\delta_k^2 J_{\theta,k}^{\top}
\bigl(v_\theta(x_k,s_k,c)-v_{\phi_i}(x_k,s_k,c)\bigr)\right],
\label{eq:local_gradient}
\end{equation}
where the expectation is over the frozen snapshot distributions in
Eq.~\ref{eq:local}, valid terms are selected by the same mask as
Eq.~\ref{eq:masked_ode_loss}, and $q_{\bar\theta}$ and $d_i^{\bar\theta}$
are not differentiated. Thus, OPD performs a local projection of the student
field on the student-induced outer and inner state distributions. The teacher
labels every valid chunk at the student's visited states, including failed
rollouts; we differentiate Eq.~\ref{eq:ode_loss} directly through the
student's VLM LoRA and action expert with stored states and teacher targets
fixed, then refresh the rollout snapshot.

\subsection{Reference-Policy Regularization for Retention}
\label{sec:reference_retention}
To preserve capabilities outside the adaptation set, we freeze the starting
policy $\pi_{\mathrm{ref}}=\pi_{\theta_0}$ and match its step means on
retained-task observations, using the same rule as the specialists.

Let $\mathcal D_{\mathrm{retain}}$ denote these retained-task observations
and $q_{\bar\theta}(x_{0:K}\mid c)$ the student snapshot's ODE chain
distribution. For valid chunks, the reference loss is
\begin{equation}
\mathcal L_{\mathrm{ref}}(\theta;\bar\theta)=
\E_{\substack{c\sim\mathcal D_{\mathrm{retain}}\\
 x_{0:K}\sim q_{\bar\theta}(\cdot\mid c)}}
\left[\frac{1}{2KHd}\sum_{k=0}^{K-1}
 \|\mu_{\theta,k}(x_k,c)
 -\operatorname{sg}[\mu_{\theta_0,k}(x_k,c)]\|_F^2\right].
\label{eq:reference_mean}
\end{equation}
The combined objective is
\begin{equation}
\mathcal L_{\mathrm{total}}=\mathcal L_{\mathrm{ODE}}
 +\beta\mathcal L_{\mathrm{ref}},\qquad\beta\geq0.
\label{eq:retention_total}
\end{equation}
We call this reference mean matching. At each retained-task state, it is the
local Gaussian-KL proxy induced by Eq.~\ref{eq:gaussian}. Retained-task
observations can come from environment resets without expert trajectories.

\section{Experiments}
\label{sec:experiments}
\subsection{Experimental Setup}
We evaluate success-once throughout, counting an episode as successful if the
task is completed at any point during execution, and report the unweighted mean when
multiple suites or tasks are available. Student consolidation updates the VLM
LoRA modules and the full action expert, while the remaining VLM parameters
and all teacher parameters are frozen.

\paragraph{LIBERO.}
We use the four single-arm manipulation suites Spatial, Object, Goal, and
LIBERO-10 from LIBERO~\citep{libero}. To test the extreme data-scarcity
setting, the student is initialized by SFT on a single trajectory per task
(1-traj), i.e., one demonstration per task. The full-data reference setting uses 50 trajectories per task. Student
checkpoint is obtained with the $\pi_{0.5}$ SFT recipe in RLinf~\citep{rlinf}.
For Spatial, Object, and Goal, the teacher is RLinf's PPO-trained multi-task
checkpoint. The LIBERO-10 teacher is obtained by further training the same
checkpoint on that suite with PPO for 120 steps. We use the standard LIBERO
success-once evaluation protocol. Each LIBERO suite is evaluated on
500 episodes; we report one training seed and compare the same four-suite
success rates across methods.

\paragraph{RoboTwin 2.0.}
We select three representative dual-arm tasks with different episode horizons
from RoboTwin 2.0~\citep{robotwin2}: \emph{Place Empty Cup},
\emph{Pick Dual Bottles}, and \emph{Stack Bowls Two}.
The common student initialization is a multi-task
SFT model trained on 50 clean trajectories per task. The ``Student
initialization'' row reports this checkpoint; VLA-MOPD further consolidates it
with task-routed PPO teachers, while the teacher row reports the specialists
themselves. Each RoboTwin task is evaluated on 100 episodes using the official
task-specific limits, and we report success-once rates.

\subsection{Main Results}
\noindent\textbf{LIBERO.}
Table~\ref{tab:libero_main} places published references alongside our
controlled runs. Published references use their reported backbones, data, and
evaluation protocols, while our method comparisons use the same initialization
and update protocol; the average is the unweighted mean of the four suites.

\begin{table}[t]
\caption{Main results on LIBERO (success-once rate, \%). VLA-MOPD is trained
from a 1-traj-per-task student initialization. Published references use their
reported backbones and data. The routed teacher is the RLinf multi-task PPO
policy for Spatial, Object, and Goal, and its LIBERO-10-adapted checkpoint for
LIBERO-10.}
\label{tab:libero_main}
\centering\small
\setlength{\tabcolsep}{3pt}
\begin{tabularx}{\linewidth}{@{}>{\raggedright\arraybackslash}Xccccc@{}}
\toprule
Method & Spatial & Object & Goal & LIBERO-10 & Avg. \\
\midrule
\multicolumn{6}{@{}l}{\textit{Full dataset (50-traj)}} \\
Octo~\citep{octo} & 78.9 & 85.7 & 84.6 & 51.1 & 75.1 \\
OpenVLA~\citep{openvla} & 84.7 & 88.4 & 79.2 & 53.7 & 76.5 \\
NORA~\citep{nora} & 92.2 & 95.4 & 89.4 & 74.6 & 87.9 \\
$\pi_0$ + FAST~\citep{pi0,pi0fast} & 96.4 & 96.8 & 88.6 & 60.2 & 85.5 \\
$\pi_{0.5}$~\citep{pi05} & 98.8 & 98.2 & 98.0 & 92.4 & 96.9 \\
\midrule
\multicolumn{6}{@{}l}{\textit{Our runs}} \\
Student initialization (base SFT, 1-traj) & 85.8 & 95.6 & 78.0 & 42.2 & 75.4 \\
Routed PPO teachers & 97.4 & \textbf{99.2} & \textbf{98.6} & 98.8 & 98.5 \\
Endpoint action distillation & 99.0 & \textbf{99.2} & 96.8 & 97.6 & 98.2 \\
\textbf{VLA-MOPD (ours)} & \textbf{99.4} & 99.0 & 98.0 & \textbf{99.0} & \textbf{98.9} \\
\bottomrule
\end{tabularx}
\end{table}

\noindent\textbf{RoboTwin 2.0.}
Table~\ref{tab:robotwin_main} reports the three-task RoboTwin results. Starting
from the common 50-traj SFT student, VLA-MOPD improves the mean success-once
rate from 73.0\% to 89.3\%, closing most of the gap to the routed teachers
(94.0\%).

\begin{table}[t]
\caption{Main results on RoboTwin 2.0 (success-once rate, \%). Each entry is
evaluated over 100 episodes.}
\label{tab:robotwin_main}
\centering
\small
\setlength{\tabcolsep}{6pt}
\begin{tabular}{@{}lcccc@{}}
\toprule
Method & \mbox{Place Empty Cup} & \mbox{Pick Dual Bottles} &
\mbox{Stack Bowls Two} & Avg. \\
\midrule
Student initialization (50-traj SFT) & 69.0 & 75.0 & 75.0 & 73.0 \\
Routed PPO teachers & \textbf{93.0} & \textbf{97.0} &
\textbf{92.0} & \textbf{94.0} \\
\textbf{VLA-MOPD (ours)} & 87.0 & 93.0 & 88.0 & 89.3 \\
\bottomrule
\end{tabular}
\end{table}

\subsection{Ablations}
\begin{figure}[t]
\centering
\includegraphics[width=\linewidth]{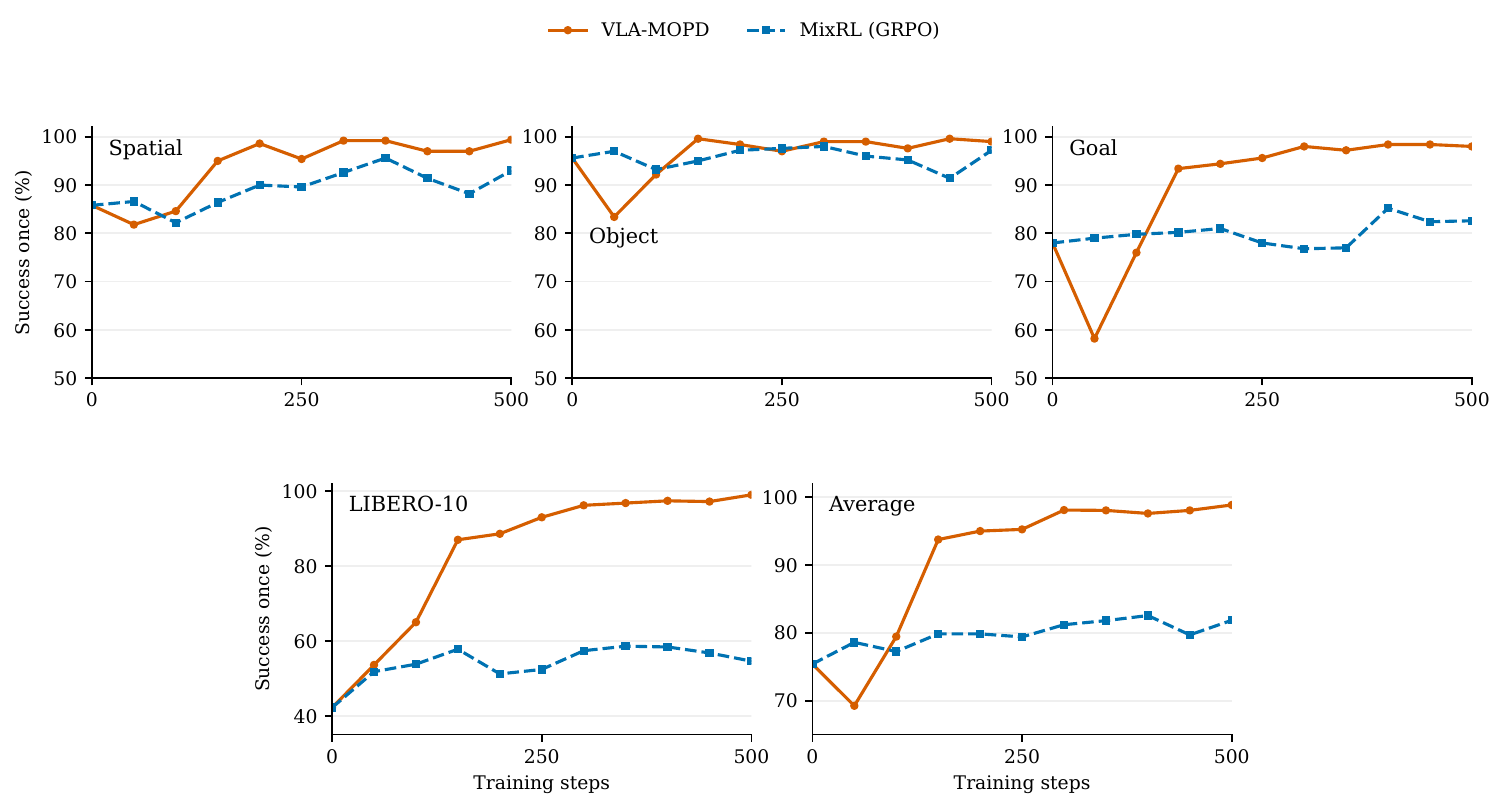}
\caption{Convergence over the first 500 training steps. The five panels show
success-once rates for the four LIBERO suites and their unweighted average.
Points mark available evaluations. All suite scores use 500 evaluation
episodes. MixRL and VLA-MOPD use matched per-step rollout budgets and are plotted at
50-step checkpoints without smoothing.}
\label{fig:convergence}
\end{figure}

\noindent\textbf{Training Efficiency.}
We compare VLA-MOPD and MixRL over the first 500
training steps, starting from the same multi-task SFT student.
Figure~\ref{fig:convergence} reports success-once rates for each LIBERO suite
and their unweighted average. We measure efficiency by convergence in student
optimizer steps under a matched run configuration. MixRL uses the GRPO
algorithm implemented as RLinf's $\pi_{\mathrm{RL}}$ baseline~\citep{pirl,rlinf}, with the same environment count and rollout count per
optimizer step as the main VLA-MOPD run; we compare the two over the first
500 steps of each run.
The recorded VLA-MOPD run improves non-monotonically from 75.4\%
at initialization to 98.1\% by step 300 and 98.85\% at the last recorded
checkpoint; the early dip is visible in Figure~\ref{fig:convergence}. MixRL
remains lower and oscillatory:
at the shared step 400, it reaches 82.55\% versus 97.6\% for VLA-MOPD. The gap is largest on Goal and
LIBERO-10. The gap is consistent with VLA-MOPD receiving teacher transition targets at
each student observation and action chunk, while MixRL must discover successful
trajectories from a terminal binary reward.

\noindent\textbf{Intermediate flow-state supervision.}
To isolate the effect of the supervision target, we compare VLA-MOPD with
endpoint action distillation under the same training setup. Both methods collect
observations through student on-policy rollouts; VLA-MOPD supervises intermediate
flow transitions, whereas endpoint action distillation supervises only the
final action. At 500 steps, endpoint action
distillation reaches 98.15\% average success, while VLA-MOPD reaches 98.9\%.
Spatial and Object are comparable (99.0\% vs. 99.4\%, and 99.2\% vs.
99.0\%, respectively); VLA-MOPD gains 1.2 points on Goal and 1.4 points on
LIBERO-10. One possible explanation is that intermediate transition supervision
helps on more compositional goals and longer-horizon tasks, where local action
errors can accumulate across action chunks.

\begin{figure}[t]
\centering
\includegraphics[width=\linewidth]{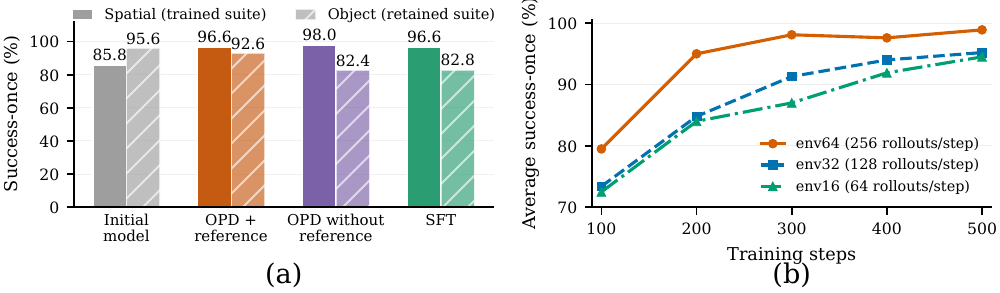}
\caption{Retention and rollout-budget analyses. (a) Object retention at
model snapshots selected for similar Spatial adaptation from checkpoints saved
at fixed step intervals. The selected Spatial scores differ by at most $1.4$
percentage points; Object success measures retention. (b) Average success during rollout-budget
sensitivity with 64, 32, and 16 parallel environments.}
\label{fig:retention_rollout}
\end{figure}

\noindent\textbf{Forgetting and reference retention.}
This experiment considers selective adaptation: stronger teachers are available
only for tasks targeted for improvement, while performance on other tasks
should be retained. We use the Spatial specialist to improve Spatial and a
frozen copy of the starting student as the reference on Object, without
requiring an additional Object specialist.
This ablation trains on Spatial and evaluates Object with reference OPD
($\beta=1$), no-reference OPD ($\beta=0$), and SFT. We define Object forgetting as the change from
the common base,
$\Delta_{\mathrm{Object}}=S_{\mathrm{Object}}(\student)-S_{\mathrm{Object}}(\base)$;
negative values indicate forgetting. From checkpoints saved at fixed step
intervals, we select snapshots with Spatial success as close as possible and
compare their Object success
(Figure~\ref{fig:retention_rollout}(a)).
From the $85.8\%$ Spatial/$95.6\%$ Object base, the selected Spatial scores of
$96.6\%$, $98.0\%$, and $96.6\%$ yield Object scores of $92.6\%$, $82.4\%$,
and $82.8\%$ for reference OPD, no-reference OPD, and SFT. The Spatial scores
differ by at most $1.4$ percentage points. The corresponding Object
drops are $3.0$, $13.2$, and $12.8$ points; reference OPD reduces the drop by
$10.2$ points versus no-reference OPD.

\noindent\textbf{Rollout-Budget Sensitivity.}
The 64-, 32-, and 16-environment settings collect 256, 128, and 64 trajectories per optimizer
step. Figure~\ref{fig:retention_rollout}(b) shows that the 64-env setting reaches $95.0\%$ at
step 200 and $98.1\%$ at step 300, compared with $84.8/91.3\%$ for the 32-env and
$84.0/87.0\%$ for the 16-env settings. The gap narrows by step 500, when the three settings
reach $98.9\%$, $95.2\%$, and $94.5\%$. These curves characterize
convergence per optimizer step under the three rollout-budget settings.

\subsection{Real-World Experiment}
We use real-world data from a Piper single-arm robot and consider three tasks:
Pick Banana, Stack Cup, and Sort Number. The student is initialized by
mixed-task SFT of a $\pi_{0.5}$ model on the collected multi-task real-world
demonstrations. For each task, we train a separate $\pi_{0.5}$ teacher with
single-task SFT on that task's real-world demonstrations. The student then
performs two rollout rounds; each round collects 50 trajectories per task. We
use the resulting student observations and saved flow states as on-policy data
for OPD, and train for one epoch after each rollout round. Over 30 evaluation
trials per task, VLA-MOPD reaches 90.0\%, compared with 70.0\% for the student
and 93.3\% for the routed teacher. Task-wise rates are shown in
Table~\ref{tab:real_world}.

\begin{figure}[H]
\centering
\includegraphics[width=\linewidth]{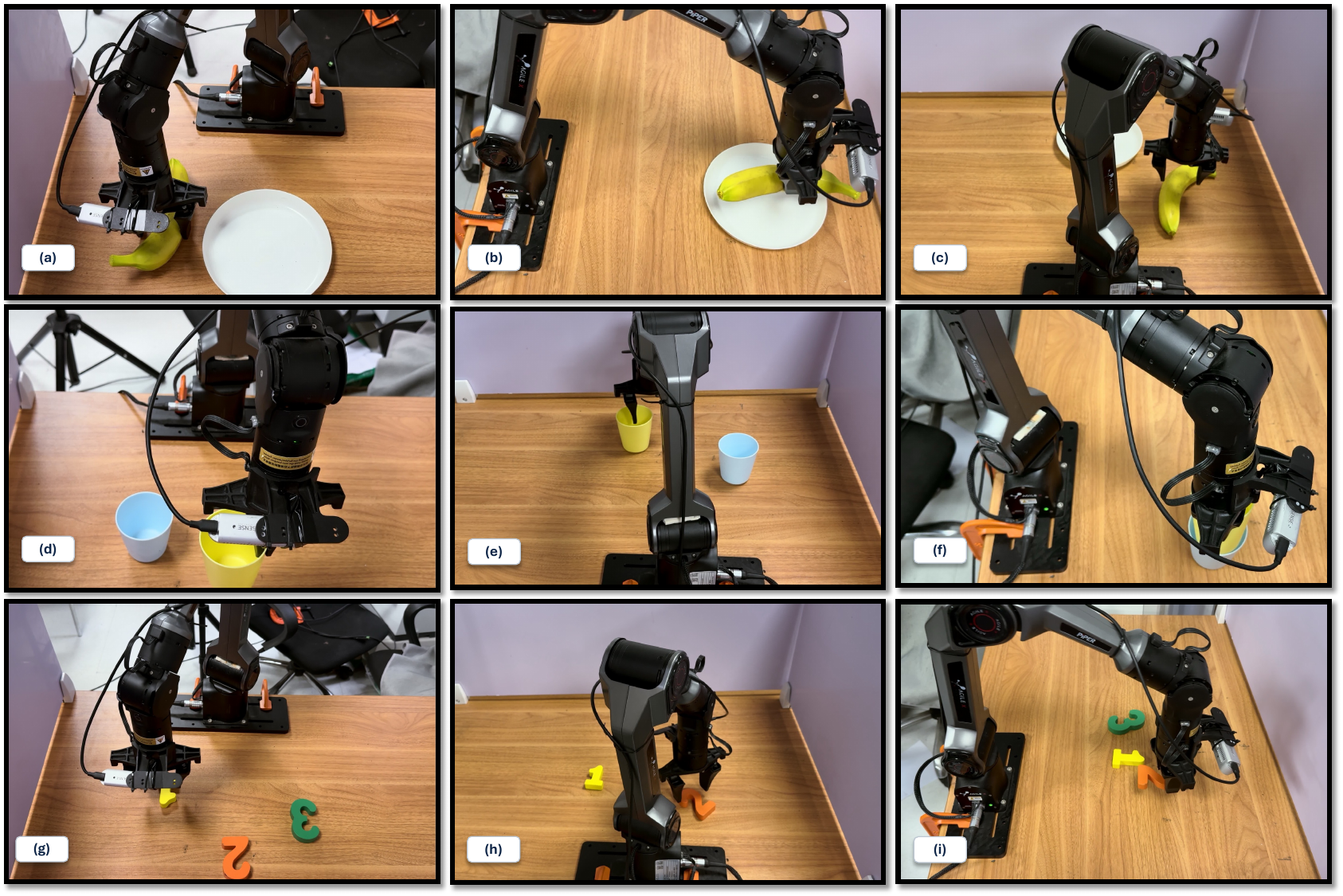}
\caption{Real-world experiment scenes. Rows show Pick Banana (a--c),
Stack Cup (d--f), and Sort Number (g--i).}
\label{fig:real_world}
\end{figure}

\begin{table}[H]
\caption{Real-world experiment success-once rates (\%; 30 trials per task).}
\label{tab:real_world}
\centering
\small
\setlength{\tabcolsep}{7pt}
\begin{tabular}{lcccc}
\toprule
Method & Pick Banana & Stack Cup & Sort Number & Avg. \\
\midrule
Student & 86.7 & 63.3 & 60.0 & 70.0 \\
Teacher & \textbf{100.0} & \textbf{96.7} & \textbf{83.3} & \textbf{93.3} \\
\textbf{VLA-MOPD} & \textbf{100.0} & 93.3 & 76.7 & 90.0 \\
\bottomrule
\end{tabular}
\end{table}

\section{Conclusion}
\label{sec:conclusion}
We present VLA-MOPD as a two-stage protocol for consolidating task or suite
specialists into one flow-matching VLA. The student follows its own rollouts
and receives local transition targets at the intermediate states of its action
generator, connecting specialist supervision to the states encountered by the
consolidated policy. Across the evaluated benchmarks and real-robot setting,
the resulting student demonstrates effective multi-task capability
consolidation and improves over its initialization.

\section{Limitations}
\label{sec:limitations}
VLA-MOPD requires task or suite specialist teachers before consolidation.
Acquiring and storing these teachers adds upfront training and maintenance
cost, especially as the number of task families grows. The quality and task
coverage of the available teachers also determine which capabilities the
student can consolidate.

\FloatBarrier
\bibliographystyle{plainnat}
\bibliography{references}
\clearpage
\beginappendix
\setcounter{section}{0}
\section{Implementation Details}
\label{app:implementation}
\begin{tabularx}{\linewidth}{@{}p{0.29\linewidth}X@{}}
\toprule
Field & Setting\\
\midrule
Base policy & $\pi_{0.5}$; multi-task SFT student initialization.\\
Student updates & Native VLM LoRA parameters and the full action expert; the remaining VLM parameters are frozen.\\
Teacher bank & Frozen teacher checkpoints routed by the stored teacher identifier.\\
Conditioning & Teacher and student encode the same observation independently and each reuse their own prefix across flow steps.\\
Rollout & Student environment interaction; saved normalized ODE states for every action chunk. The evaluated flow configuration uses five solver steps and the same step schedule for teacher and student targets.\\
Action horizon & LIBERO action chunks contain 10 actions; RoboTwin uses an action horizon of 50.\\
Objective & Half squared ODE step-mean error, averaged over flow steps and valid model action elements.\\
Teacher targets & Computed on stored student states and cached before student optimization.\\
Backward pass & Shared trainable prefix graph; losses from all selected flow steps are averaged before backward.\\
Rollout budget & 64 training environments; four rollout epochs per optimizer step, one for each suite (Spatial/Object/Goal/LIBERO-10). The suite-specific maximum steps are 240/240/320/480, respectively.\\
Actor batches & Microbatch size 8 and global batch size 1024.\\
Optimization & Adam with learning rate $2.5\times10^{-5}$, $(\beta_1,\beta_2)=(0.9,0.95)$, weight decay $0.01$, gradient clipping at 1.0, and cosine decay.\\
Training schedule & 1000 optimizer steps; results are reported over the first 500 steps. Checkpoints are saved every 50 steps; random seed 42.\\
OPD target computation & Teacher microbatch size 4; all five flow steps are used; $\beta=0$ for the main consolidation experiments, with the teacher-ratio schedule fixed at zero. The forgetting ablation compares reference matching ($\beta=1$) with no reference matching ($\beta=0$).\\
\bottomrule
\end{tabularx}

The implementation uses the configured five-step \texttt{flow\_ode} schedule; the cached teacher target and student prediction use the same normalized step coordinates. The action dimension $d$ in Eq.~\ref{eq:ode_loss} denotes the model’s internal
normalized action representation, which can exceed the number of robot
control coordinates. Matching is performed before decoding and uses all
coordinates returned by the action head. Normalization assets are shared
between teacher and student. This aligns their output coordinates while
allowing their VLM encoders and action experts to have different weights.

For an update microbatch $\mathcal B$, let $m_{b,h}\in\{0,1\}$ be the
validity mask, broadcast over the $d$ model action coordinates. The estimator
of Eq.~\ref{eq:ode_loss} used by the implementation is
\begin{equation}
\widehat{\mathcal L}_{\mathrm{ODE}}
=\frac{1}{2K Z_{\mathcal B}}
\sum_{b\in\mathcal B}\sum_{k=0}^{K-1}\sum_{h=1}^{H}\sum_{j=1}^{d}
 m_{b,h}\bigl[\mu_{\theta,b,k,h,j}-\widehat\mu_{b,k,h,j}\bigr]^2,
\label{eq:masked_ode_loss}
\end{equation}
where $Z_{\mathcal B}=\max(1,d\sum_{b,h}m_{b,h})$,
$\mu_{\theta,b,k}=\mu_{\theta,k}(x_{b,k},c_b)$, and
$\widehat\mu_{b,k}=\operatorname{sg}[\mu_{\phi_{i_b},k}(x_{b,k},c_b)]$.
For chunk-level validity masks, the mask is broadcast over all positions and
coordinates of the action chunk. Completed rollout slots and padded samples
are excluded. The implementation computes a normalized masked loss within
each microbatch, then averages its gradients across accumulation steps and
distributed workers. Accordingly, it averages microbatch-normalized losses;
when valid counts differ, this need not equal one global valid-element mean.
Both successful and unsuccessful episodes can supply valid training samples;
restricting distillation to failed episodes is an optional sampling filter.
The student computes its trainable VLM prefix once per microbatch and reuses
it across the denoising-step predictions. Their losses are averaged before a
joint backward pass through the shared prefix graph and action expert.

\section{From Reverse KL to ODE Step-Mean Matching}
\label{app:reverse_kl}
This section makes explicit the connection between the reverse-KL definition
in DiffusionOPD \citep{diffusionopd} and the deterministic loss used in our
implementation. Fix an instruction--observation pair $c$, a stored student
state $x_k$, and flow step $k$. Let
$p_{\theta,k}(\cdot\mid x_k,c)$ denote the student transition and let
$p^{\mathrm T}_{i,k}(\cdot\mid x_k,c)$ denote the transition of the routed,
frozen teacher. The reverse transition objective is
\begin{equation}
\mathcal L_{\mathrm{rev}}
=\E_{\substack{i\sim p(i),\ c\sim d_i^{\bar\theta}\\
 x_{0:K}\sim q_{\bar\theta}(\cdot\mid c)}}
\left[\frac{1}{K}\sum_{k=0}^{K-1}
 \KL\!\left(p_{\theta,k}(\cdot\mid x_k,c)\,\middle\|\,
 p^{\mathrm T}_{i,k}(\cdot\mid x_k,c)\right)\right].
\label{eq:app_reverse_kl}
\end{equation}
The first argument is the student, so this is the reverse KL convention used
here. The snapshot distribution $d_i^{\bar\theta}$ and the stored-state law
$q_{\bar\theta}$ are held fixed during the update.

\paragraph{Reverse KL for shared-covariance transitions.}
Assume temporarily that both transitions are Gaussian,
$p_{\theta,k}=\mathcal N(\mu_{\theta,k},\Sigma_k)$ and
$p^{\mathrm T}_{i,k}=\mathcal N(\mu^{\mathrm T}_{i,k},\Sigma_k)$, with the
same positive covariance at the same state. The general Gaussian KL reduces to
\begin{equation}
\KL(p_{\theta,k}\|p^{\mathrm T}_{i,k})
=\frac{1}{2}
 (\mu_{\theta,k}-\mu^{\mathrm T}_{i,k})^{\!\top}
 \Sigma_k^{-1}
 (\mu_{\theta,k}-\mu^{\mathrm T}_{i,k}).
\label{eq:app_shared_cov_kl}
\end{equation}
For $\Sigma_k=\sigma_k^2I$, this is
\begin{equation}
\KL(p_{\theta,k}\|p^{\mathrm T}_{i,k})
=\frac{1}{2\sigma_k^2}
 \left\|\mu_{\theta,k}-\mu^{\mathrm T}_{i,k}\right\|_F^2.
\label{eq:app_mean_kl}
\end{equation}

\paragraph{Velocity-field form.}
Write one Euler transition of the student and teacher flow fields as
\begin{equation}
\mu_{\theta,k}=x_k-\delta_k v_\theta(x_k,s_k,c),\qquad
\mu^{\mathrm T}_{i,k}=x_k-\delta_k v^{\mathrm T}_{i}(x_k,s_k,c).
\label{eq:app_euler_means}
\end{equation}
Because both transitions start from the same stored state, the state cancels:
\begin{equation}
\mu_{\theta,k}-\mu^{\mathrm T}_{i,k}
=-\delta_k\bigl(v_\theta(x_k,s_k,c)-v^{\mathrm T}_{i}(x_k,s_k,c)\bigr).
\label{eq:app_mean_velocity_identity}
\end{equation}
Substituting Eq.~\ref{eq:app_mean_velocity_identity} into
Eq.~\ref{eq:app_mean_kl} gives the step-size-weighted velocity objective
\begin{equation}
\KL(p_{\theta,k}\|p^{\mathrm T}_{i,k})
=\frac{\delta_k^2}{2\sigma_k^2}
 \left\|v_\theta(x_k,s_k,c)-v^{\mathrm T}_{i}(x_k,s_k,c)\right\|_F^2.
\label{eq:app_velocity_kl}
\end{equation}
Thus the reverse-KL transition view and velocity-field regression are the same
quadratic form under the shared-covariance assumption.

\paragraph{Deterministic ODE transition matching.}
Following the deterministic regime in DiffusionOPD, our sampler uses direct
squared matching of ODE step means rather than a conditional KL:
\begin{equation}
\mathcal L_{\mathrm{mean}}
=\E_{\substack{i\sim p(i),\ c\sim d_i^{\bar\theta}\\
 x_{0:K}\sim q_{\bar\theta}(\cdot\mid c)}}
\left[\frac{1}{2KHd}\sum_{k=0}^{K-1}
 \left\|\mu_{\theta,k}(x_k,c)
 -\operatorname{sg}\!\left[\mu^{\mathrm T}_{i,k}(x_k,c)\right]\right\|_F^2\right].
\label{eq:app_mean_loss}
\end{equation}
The stop-gradient keeps the teacher target fixed. Combining
Eq.~\ref{eq:app_mean_velocity_identity} with Eq.~\ref{eq:app_mean_loss}
shows that each term is proportional to
$\delta_k^2\|v_\theta-v_i^{\mathrm T}\|_F^2$; the factor $1/(Hd)$ is only the
per-element normalization. If the shared covariance is constant, the omitted
$\sigma^{-2}$ is a global scale absorbed by the loss weight or learning rate.
If $\sigma_k$ varies with $k$, using Eq.~\ref{eq:app_mean_loss} amounts to an
explicit choice of uniform mean-space weighting rather than the additional
$\sigma_k^{-2}$ time weighting in Eq.~\ref{eq:app_velocity_kl}.

The deterministic loss is the ODE mean-matching counterpart of the stochastic
transition objective. It is not obtained by setting
$\sigma_k^2=0$ in a Gaussian KL: distinct Dirac transitions do not have a
finite KL. Nor does Eq.~\ref{eq:app_mean_loss} claim to be an exact KL between
the final action distributions. Its on-policy character comes from evaluating
both fields at the saved student states $x_k$; the recorded rollout law is
detached, while gradients flow through the current student's prefix and action
expert. This is also why the teacher is queried at every stored state rather
than used to generate a separate complete action chunk.

\section{Real-World Evaluation Details}
\label{app:real_world}
\paragraph{Hardware and observations.}
We use a single-arm Piper robot with six arm joints and a gripper.
The policy receives red-green-blue (RGB) observations from a main-view camera and a
wrist-mounted camera. Images are recorded at a resolution of
$640\times480$ and resized with aspect-ratio-preserving padding to
$224\times224$ for model input. The nominal action-execution frequency
is 30 Hz.

\paragraph{Action representation.}
The physical action space consists of six joint-position targets and
one gripper-opening target. The arm-joint targets are represented as
offsets from the current joint positions, whereas the gripper target
remains absolute. Before execution, the joint offsets are converted
back to absolute joint-position targets. The model uses a
32-dimensional internal action representation, of which the first
seven dimensions are used for Piper control.

\paragraph{Demonstrations and specialist training.}
The student is initialized by mixed-task SFT on demonstrations from
Pick Banana, Stack Cup, and Sort Number. Each specialist teacher is
trained separately by single-task SFT on demonstrations from its
corresponding task.

\paragraph{On-policy training and evaluation.}
For OPD training, the student performs two rollout rounds. Each round collects
50 trajectories per task, and the resulting on-policy data is used for one
training epoch before the next rollout round. The training policy predicts and
executes action horizons of 50, with five flow-ODE denoising steps.
For the Piper evaluation, the policy predicts an action horizon of 50 but the
controller executes only the first 25 actions before requesting the next
prediction.
\begin{table}[H]
\centering
\small
\begin{tabularx}{\linewidth}{@{}lX@{}}
\toprule
Task & Instruction \\
\midrule
Pick Banana & ``Grasp the banana from the table and place it on the plate.'' \\
Stack Cup & ``Stack the yellow cup into the blue cup.'' \\
Sort Number & ``Sort the numbers.'' \\
\bottomrule
\end{tabularx}
\end{table}

\end{document}